\documentclass[conference,10pt]{IEEEtran}
\IEEEoverridecommandlockouts
\usepackage{cite}
\usepackage{amsmath,amssymb,amsfonts}
\usepackage{algorithmic}
\usepackage{graphicx}
\usepackage{textcomp}
\usepackage{xcolor}

\usepackage{subcaption}
\usepackage[ruled]{algorithm2e}
\usepackage{caption}
\usepackage{color}
\usepackage{ctable}
\usepackage{siunitx}
\usepackage{enumitem}
\usepackage{multicol}
\usepackage{multirow}
\usepackage{setspace}
\usepackage{microtype}
\usepackage{balance}

\SetAlFnt{\small}
\SetKwInOut{Input}{Input}
\SetKwInOut{Output}{Output}
\def\BibTeX{{\rm B\kern-.05em{\sc i\kern-.025em b}\kern-.08em
    T\kern-.1667em\lower.7ex\hbox{E}\kern-.125emX}}
\begin{document}
\newcommand{\jh}{blue}

\title{Energy-Efficient Inference Accelerator for Memory-Augmented Neural Networks on an FPGA}

\author{
\IEEEauthorblockN{Seongsik Park, Jaehee Jang, Seijoon Kim, Sungroh Yoon}
\IEEEauthorblockA{
Electrical and Computer Engineering, Seoul National University, Seoul, Korea
}
{sryoon@snu.ac.kr}
}

\maketitle

\begin{abstract}
Memory-augmented neural networks (MANNs) are designed for question-answering tasks.
It is difficult to run a MANN effectively on accelerators designed for other neural networks (NNs), in particular on mobile devices, because MANNs require recurrent data paths and various types of operations related to external memory access.
We implement an accelerator for MANNs on a field-programmable gate array (FPGA) based on a data flow architecture.
Inference times are also reduced by inference thresholding, which is a data-based maximum inner-product search specialized for natural language tasks.
Measurements on the bAbI data show that the energy efficiency of the accelerator (FLOPS/kJ) was higher than that of an NVIDIA TITAN V GPU by a factor of about 125, increasing to 140 with inference thresholding.

\end{abstract}

\begin{IEEEkeywords}
deep learning, memory-augmented neural networks, inference accelerator, FPGA, data-based maximum-inner product search, question and answer
\end{IEEEkeywords}

\section{Introduction}
Deep neural networks (DNNs) require more computing power and storage than most mobile devices can provide.
So mobile DNNs are commonly trained and run on remote servers.
This limits performance, relies on network availability, and increases maintenance.
It motivates the development of on-device inference.

In a dataflow architecture (DFA), data goes directly from one processing element to another, reducing the need for energy-consuming memory accesses~\cite{horowitz20141}.
Layer-wise parallelization and recurrent paths can be implemented on DFAs, through the use of fine-grained parallelism.
DFAs have therefore been used to realize inference on mobile devices~\cite{han2016eie,rybalkin2017hardware,han2017ese}.

Memory-augmented neural networks (MANNs), which include memory networks~\cite{sukhbaatar2015end}, are recurrent neural networks (RNNs) with external memory to increase learning capacity.
MANNs require both recursive and memory operations in each layer, making them difficult to parallelize on CPUs or GPUs.

We propose an accelerator for MANNs based on a field-programmable gate array (FPGA), which uses a DFA to realize energy-efficient inference in the domain of natural language processing (NLP), which is a major application of MANNs.
We also introduce a data-based method of maximum inner-product search (MIPS), called inference thresholding, together with an efficient index ordering.
This speeds up inference and the operation time of the output layer, which is particularly important in tasks with large classes, such as NLP.

Our implementation outperformed a GPU in terms of energy efficiency (FLOPS/kJ) by a factor of 126 on the bAbI dataset~\cite{weston2015towards}, and by 140 when inference thresholding was also used.
The contributions of this paper are as follows:
\begin{itemize}[topsep=0pt,itemsep=0ex,partopsep=1ex,parsep=1ex,leftmargin=*]
    \item A streaming-based inference architecture for MANNs, which we believe is the first. 
    \item Fast inference on this hardware using inference thresholding.
    \item Implementation and validation of this approach on an FPGA.
\end{itemize}

\begin{figure*}[t]
    \centering
    \includegraphics[width=1.0\linewidth]{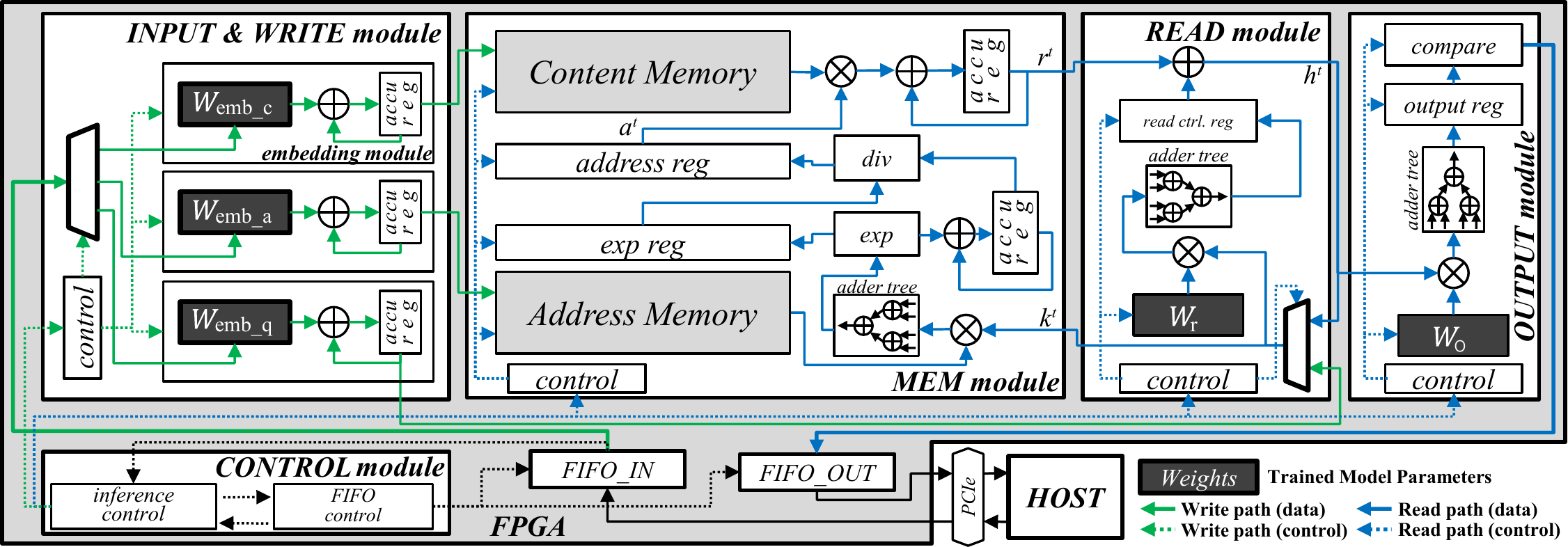}
    \vspace{-1.5em}
    \caption{Proposed architecture of the FPGA-based accelerator for MANNs}
    \label{fig:proposed_archi}
    \vspace{-1.5em}
\end{figure*}

\section{Memory-Augmented Neural Networks}
MANNs, which are RNNs with more storage, are designed for question answering (QA) and other NLP tasks~\cite{sukhbaatar2015end}.
A MANN consists of external memory and a controller, and it learns how to read and write information from and to the memory.
The memory operations of a MANN can be divided into three types: addressing, write, and read.
Content-based addressing is usually employed in MANNs, and can be expressed as follows:
\begin{equation}
\label{eq:mem_addr}
    a_{i}^{t} = \frac{\textrm{exp}\{M_{\textrm{a},i}\cdot k^{t}\}}{\sum_{j}^{L}{\textrm{exp}\{M_{\textrm{a},j}\cdot k^{t}\}}} \textrm{,}
\end{equation}
where $a_{i}^{t}$ is the read weight of the $i$th memory element at time $t$, $M_{a}$ is the address memory, $L$ is the number of memory elements, and $k^{t}$ is a read key.

Each memory element stores an embedded sentence vector as follows:
\begin{equation}
\label{eq:mem_write}
    M_{i} = W_{\textrm{emb}} S_{i} = \sum\nolimits_{\mathrm{idx} \in S_{i}}{W_{{\mathrm{emb}}_{\mathrm{:,idx}}}}\textrm{,}
\end{equation}
where $W_{\textrm{emb}}$ is a word-embedding weight, and $S_{i}$ is an input sentence consisting of word indices.
A memory read begins with the generation of a read key in the memory controller after the previous write.
The read key $k^{t}$ at time $t$ is found as follows:
\begin{equation}
\label{eq:mem_read_key}
    k^{t} =
    \begin{cases}
        W_{\textrm{emb\_q}} q  & \textrm{if}~t=1 \\
        h^{t\textrm{-}1}  & \textrm{otherwise} \textrm{,}\\
    \end{cases}
\end{equation}
where $q$ is a question vector, and $h$ is an output vector from the controller, which is described as follows:
\begin{equation}
    h^{t} = r^{t} + W_{\textrm{r}} k^{t} \textrm{,}
\end{equation}
where $r$ is a read vector, and $W_{\textrm{r}}$ is the weight of the controller.
The read vector for content-based addressing is generated by a content memory as follows: 
\begin{equation}
\label{eq:mem_read_vec}
    r^t = M_{\textrm{c}} a^t \textrm{.}
\end{equation}
The predicted label $\tilde{y}$, produced by inference is given by
\begin{equation}
\label{eq:output}
    \tilde{y} = \operatorname{arg\,max}_{i} (z_i) = \operatorname{arg\,max}_{i} (W_{{\mathrm{o}}_{i,:}} h^t) \textrm{,}
\end{equation}
where $W_{\textrm{o}}$ is the weight of the output layer, and $z_i$ is a logit with index $i$.

\section{Hardware Architecture}
Fig. 1 shows the architecture and data flow of our accelerator, which consists of several modules which receive inference data and trained models ($W_{\mathrm{emb}}$, $W_{\mathrm{r}}$, and $W_{\mathrm{o}}$) from a host computer in the form of streams through a FIFO queue.
A pre-trained model with appropriate data is passed to each module.

Control signals from the host, embedded in the data, pass to the CONTROL module, which has an inference control component that signals other modules.
For example, in a QA task, context data in the form of sentences $S$, together with the question $q$, arrive in the input stream (green line in Fig. 1).
When this stream is finished, the READ module generates a read key $k^{t}$, and the MEM module uses this key to read a vector $r^{t}$ from the context memory.
Reads can be recursive because the READ module is composed of an RNN.
After all read operations are complete, the OUTPUT module returns the answer to the question through the FIFO queue to the host.
 
The INPUT \& WRITE modules receive input data from the host and write embedded vectors to context and address memory in the MEM module.
In an NLP task, a discrete and sparse sentence vector (e.g. a bag-of-words) is converted into a dense embedded vector by the embedding layer.
If the input to a MANN includes word indices, then the efficiency of embedding in the INPUT \& WRITE module can be improved, as shown in Eq.~\ref{eq:mem_write}.
The embedding module in the INPUT \& WRITE module only needs to read the columns of the embedding weight $W_{\mathrm{emb}}$ corresponding to the indices of the input words.
This reduces the number of memory accesses needed to read the embedding weights, and the number of multiplications needed to calculate the embedding vector, which lead to improving energy efficiency.

The MEM module consists of the address memory, which is content-addressible (Eq.~\ref{eq:mem_addr}) and context memory, which generates a read vector $r^{t}$ by soft-addressing based on the attention at obtained from the address memory (Eq.~\ref{eq:mem_read_vec}).
The address and context memory together store the embedded vector from the INPUT \& WRITE module. This requires costly operations such as softmax, which incurs an exponentiation and a division, which cannot be parallelized on an FPGA.
The MEM module is therefore implemented with element-wise sequential operations which can exploit fine-grained parallelism.

The READ module is an RNN, and the OUTPUT module is a fully connected neural network.
The READ module generates the read key $k^t$ which is used to calculate the attention at in the MEM module, and receives a read vector $r^t$ from the MEM module (Eqs.~\ref{eq:mem_read_key} and \ref{eq:mem_read_vec}).
The blue line in the READ module in Fig.~\ref{fig:proposed_archi} shows how a recurrent READ path can be implemented efficiently.
 
The OUTPUT module predicts the label $\tilde{y}$ based on the read vector, by multiplying the vector $h^{t}$ and the weight matrix of the output layer $W_\mathrm{o}$, as shown in Eq.~\ref{eq:output}.
Matrix multiplication is implemented as a series of dot products because the hardware is insufficient to parallelize it directly.
In the OUTPUT module the logit $z_{i}$ of each index is sequentially calculated to find the maximum logit; this takes up a lot of the inference time.

\section{Fast Inference Method}

\subsection{Inference Thresholding}
\SetStartEndCondition{ }{}{}%
\SetKwProg{Fn}{def}{\string:}{}
\SetKwFunction{Range}{range}
\SetKw{KwTo}{in}\SetKwFor{For}{for}{\string:}{}%
\SetKwIF{If}{ElseIf}{Else}{if}{:}{elif}{else:}{}%
\SetKwFor{While}{while}{:}{fintq}%
\AlgoDontDisplayBlockMarkers\SetAlgoNoEnd\SetAlgoNoLine%
\begin{algorithm}[t]
\caption{Inference Thresholding}
\label{alg:pseudocode_infer_th}
\Input{training dataset $\mathcal{D}=\{x_n, y_n\}^{N}_{n=1}$, \\
        inference data $\tilde{x}$}
\Output{prediction label $\tilde{y}$}
\vspace{0.2em}
\textbf{Notations: } 
$z$: vector of logits, $z_{i}$: logit value at $i$th index, \\
$M$: pre-trained model, $I$: dimension of output vector, \\
$\rho$: thresholding constant, \\
$HG_{i}$: histogram of $z_{i}$ when $i = \operatorname{arg\,max}_{i} z_{i}$, \\ 
$HG_{\bar{i}}$: histogram of $z_{i}$ when $i \neq \operatorname{arg\,max}_{i} z_{i}$
\hrule
\vspace{0.2em}
\textbf{Step 1: Estimate logit distributions} \\

\For{$(x_{n}, y_{n})$ \KwTo $\mathcal{D}$}{
    $z \leftarrow $ Do forward pass $M(x_{n})$, $y \leftarrow \operatorname{arg\,max}_{i} z_{i}$\\
    \If{$y$ == $y_{n}$}{
        \For{$\mathrm{i}$ \KwTo $\mathrm{1:I}$} {
            \eIf{$i$ == $y$} {
                Update $HG_{i} \leftarrow z_{i}$
            }
            {
                Update $HG_{\bar{i}} \leftarrow z_{i}$
            }
        }
        
    }
}

\For{$i$ \KwTo $\mathrm{1:I}$} {
    Estimate $p(z_{i} \vert y = i)$ from $HG_{i}$ \\
}
\vspace{0.2em}
\hrule
\vspace{0.2em}
\textbf{Step 2: Set the inference thresholds}\\
$p(y=i \vert z_{i}) \leftarrow p(z_{i} \vert y=i) p(y=i) $\\ 
\For{$i$ \KwTo $\mathrm{1:I}$}{
    $\theta_{i} \leftarrow \operatorname{min}(\{z_{i} \vert p(y=i \vert z_{i}) \geq \rho\})$\\
}
\vspace{0.2em}
\hrule
\vspace{0.2em}
\textbf{Step 3: Set the efficient index order}\\
\For{$i$ \KwTo $\mathrm{1:I}$}{
    $S_{i} \leftarrow $ avg. silhouette coefficient of $HG_{i}$ \\
}
$A \leftarrow $ indices sorted by $S_{\mathrm{i}}$ in descending order

\vspace{0.2em}
\hrule
\vspace{0.2em}
\textbf{Step 4: Inference thresholding}\\
$h \leftarrow$ Do forward pass $M(\tilde{x})$ until output layer\\

\For{$\mathrm{i}$ \KwTo $\mathrm{1:I}$}{
    $a \leftarrow A_{i}$ \\
    $z_{a} \leftarrow W_{{\mathrm{o}}_{a,:}} h$ \\
    \If{$z_{a} > \theta_{a}$} {
        \textbf{return} $\tilde{y} \leftarrow a$ \\
    }
}

\textbf{return} $\tilde{y} \leftarrow \operatorname{arg\,max}_{i} z_{i}$

%
\end{algorithm}

\begin{figure}[t]
    \centering
    \includegraphics[width=1.0\linewidth]{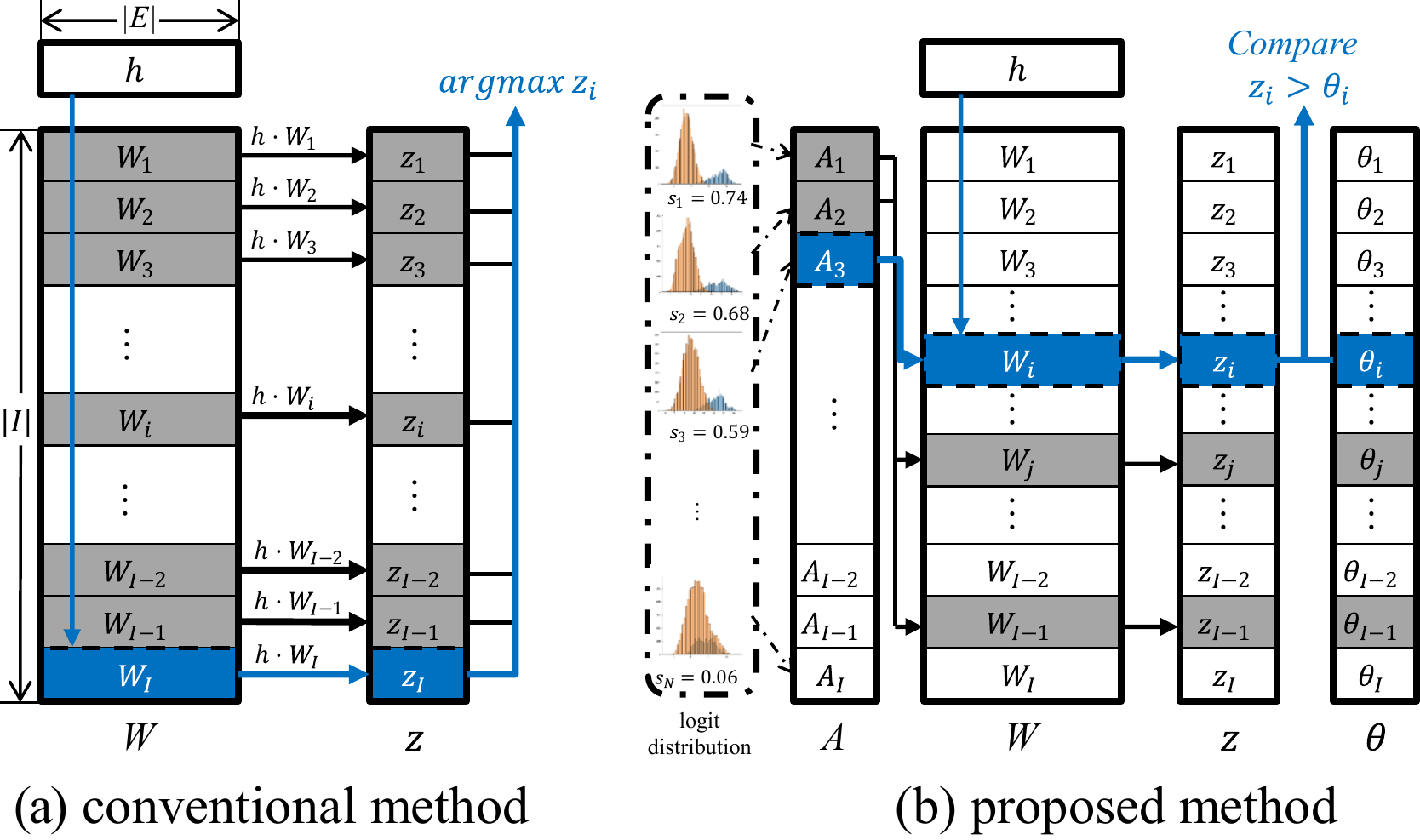}
    \vspace{-1.5em}
    \caption{
        MIPS in the OUTPUT module: (a) the conventional method needs to compare all logits; (b) inference thresholding stops the comparison if $z_i > \theta_i$.
    }
    \label{fig:output_layer}
    \vspace{-0.5em}
\end{figure}

A MANN implemented as a DFA can exploit fine-grained parallelism in each layer.
However, in an NLP task the dimension of the output $|I|$ is much larger than that of the embedding $|E|$, making it difficult to parallelize operations in the output layer~\cite{li2015fpga}.
Thus, when calculating a logit $z$ in the output layer, we must sequentially calculate the dot product of the input vector $h$ and the row of the weight matrix corresponding to the index $W_{{\mathrm{o}}{i,:}}$ in the output module, as shown in Fig.~\ref{fig:output_layer}-(a).
Because the operation time of the output layer is $O(|I|)$, the inference time increases with $|I|$.

We implement the output layer sequentially, but limit the computation required by introducing inference thresholding (Algo.~\ref{alg:pseudocode_infer_th}).
We approximate the MIPS by speculating that, given $z_{i}$, the index $i$ will be the predicted label $\tilde{y}$.
If we can conjecture the maximum logit for index $i$ with sufficient confidence, then we need not compare the remaining logits.

Inference thresholding was motivated by observing logit distributions in a trained model in which the logits $z$ are fitted to the mixture models, as shown in Fig.~\ref{fig:output_layer}-(b).
To predict whether logit $z_{i}$ is the maximum value of all logits $z$, we consider two distributions: in one, $z_{i}$ is the maximum, and in the other it is not. 

On this basis we can estimate conditional probability density functions (PDFs) $p(z_{i}|y = i)$ for the training label $y$ by kernel density estimation (Step 1 in Algo.~\ref{alg:pseudocode_infer_th}).
The PDFs of the inference dataset can be approximated by those obtained from the training dataset. 
By applying Bayes' theorem to the approximated PDFs, we can obtain the posteriors of the logits for the inference dataset $p(\tilde{y}=i \vert z_{i})$ as follows:
\begin{equation}
    \label{eq:bayes}
    p(\tilde{y}=i \vert z_{i}) \approx p(y=i \vert z_{i}) \propto p(z_{i} \vert y=i) p(y=i) \textrm{,}
\end{equation}
where $P(y=i)$ is the probability that the index $i$ is a training label $y$.

To apply estimated probabilities to the inference process in the output layer, we compare each logit $z_{i}$ with a threshold $\theta_{i}$, which is the the smallest value of those logits of which the estimated posterior probability $p(y = i|z_{i})$ is larger than $\rho$: 
\begin{equation}
\theta_{i} := \mathrm{min}(\{z_{i}|p(y = i|z_{i}) \geq \rho\}) \textrm{,}
\end{equation}
where $\rho$ is a thresholding constant (Step 2 in Algo.~\ref{alg:pseudocode_infer_th}). This yields a speculative value for the label.

\begin{figure}[t]
    \centering
    \includegraphics[width=1.0\linewidth]{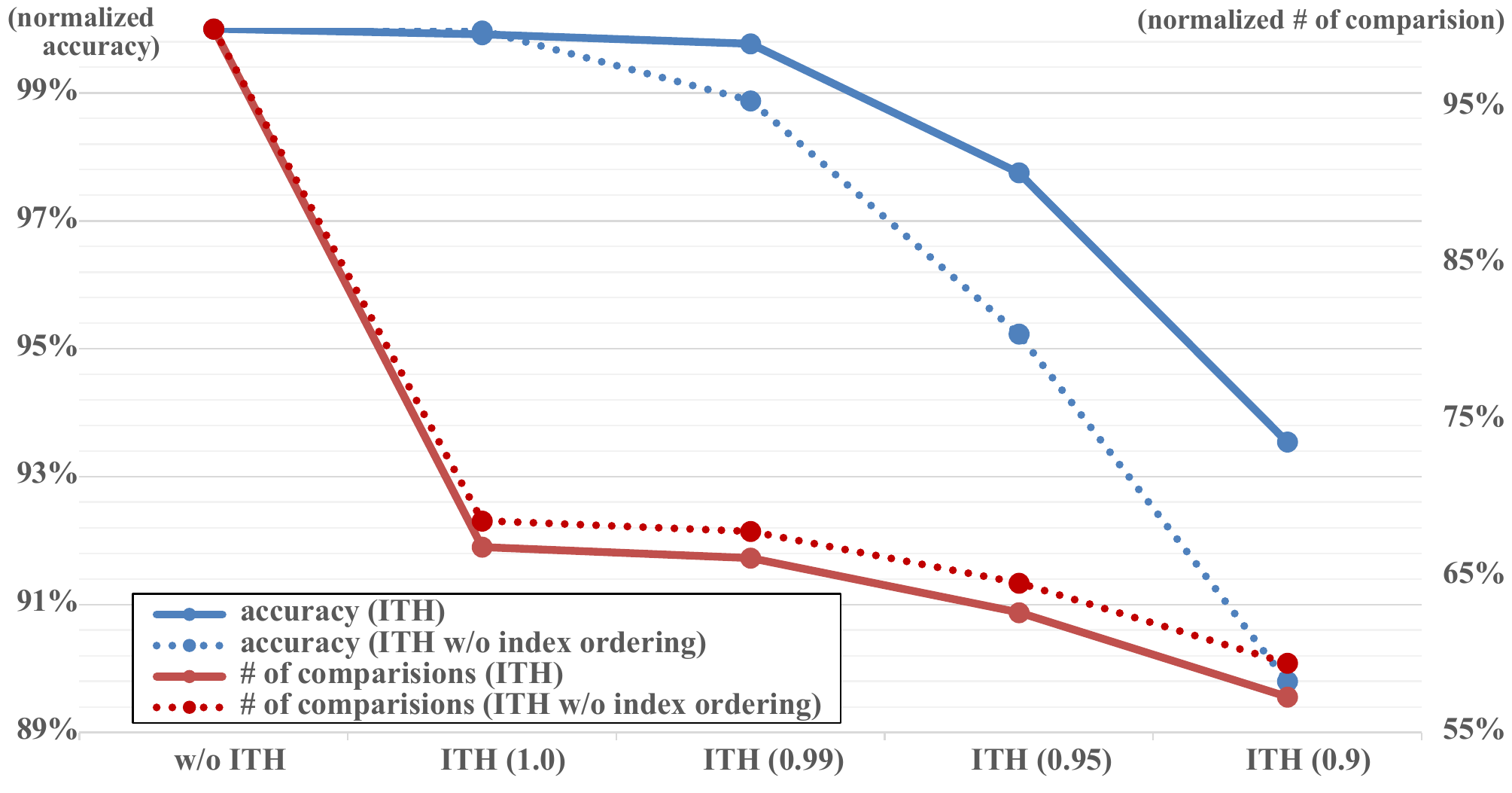}
    \vspace{-1.5em}
    \caption{Evaluation of the effect of inference thresholding and index ordering: in terms of the accuracy and number of comparisons required in the MIPS against threshold constant $\rho$, on the bAbI dataset (ITH = inference thresholding).
    }
    \label{fig:infer_th}
    \vspace{-0.5em}
\end{figure}

\subsection{Efficient Index Order for Inference Thresholding}
Inference thresholding is quicker if we order the logits so that those for which thresholding is most effective come first (Fig.~\ref{fig:output_layer}).
This can be seen as determining whether the logit belongs to the class $y = i$.
From this perspective, inference thresholding will be more effective for a logit with a long inter-class distance and a short intra-class distance.
We therefore sort the indices into descending order of silhouette coefficient~\cite{rousseeuw1987silhouettes} (Step 3 in Algo.~\ref{alg:pseudocode_infer_th}).

The effect of inference thresholding and index ordering is depicted in Fig.~\ref{fig:infer_th}. As the threshold constant $\rho$ decreases, MIPS requires fewer comparisons but accuracy declines. Ordering improves both accuracy and speed.

%


\section{Experimental Results}
%
%
\begin{figure*}[t]
    \includegraphics[width=1.0\linewidth]{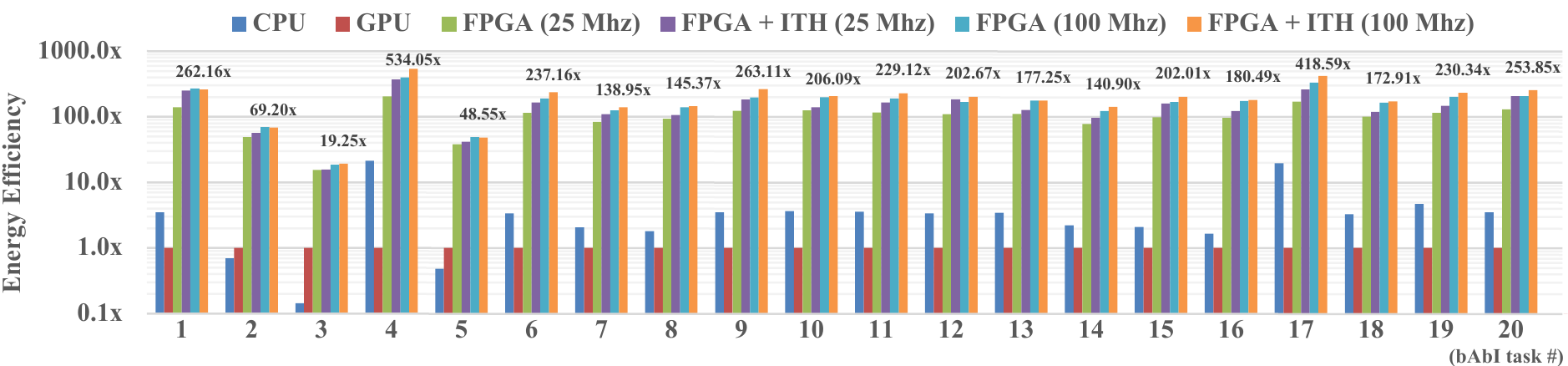}
    \vspace{-1.5em}
    \caption{\textls[-40]{Energy efficiency of inference on the bAbI dataset on various configurations compared with the GPU (ITH = inference thresholding).}}
    \label{fig:result_efficiency}
    \vspace{-1.5em}
\end{figure*}

%

We implemented the accelerator and measured the inference time and power consumption on an Intel Core i9-7900X CPU, and on an NVIDIA TITAN V GPU, and a Xilinx Virtex UltraScale VCU107 FPGA linked to the same CPU.

Time and power measurement were made for 20 tasks from the bAbI QA dataset~\cite{weston2015towards}.
Timings, which included transmission of the pre-trained model and inference data to the GPU and FPGA, were repeated 100 times; power measurements were made over five minutes.
We ran the FPGA at 25, 50, 75, and 100 MHz to evaluate the effect of the host-FPGA interface.
We set the thresholding constant $\rho$ to 1.0, which reduced accuracy by less than 0.1\%.

Averaged timings and power measurements are listed in Table~\ref{tab:experiment_result_gain}.
Running on the FPGA, the accelerator took less time at higher frequencies, as we would expect: but the improvement was not linear.
Inference thresholding reduced timings by 6-18\%, depending on frequency.
The accelerator ran between 5.2 and 7.5 times faster than the GPU, and between 5.6 and 8.0 times faster than the CPU.
The GPU used most power, and the FPGA running at 25MHz used least.
The CPU used 1.7 times less energy than the GPU, and the FPGA used 74 times less, or 140 times less using inference thresholding. 

Results on individual tasks are shown in Fig.~\ref{fig:result_efficiency}, again normalized to the performance of the GPU.
The FPGA implementation was the most energy-efficient across all tasks, and inference thresholding increased the margin.

\ctable[
pos = t,
center,
caption = {Average measurement results, speedup, and energy-efficiency of inference on the bAbI dataset},
captionskip = -1ex,
mincapwidth = \columnwidth,
label = {tab:experiment_result_gain},
doinside = {\footnotesize \def\arraystretch{.7}}
]{l|rr|rr}{
    \tnote[a]{normalized to the result on the GPU}
}{
    \toprule
    Configurations & Time (s) & Power (W) & Speedup\tmark[a] & FLOPS/kJ\tmark[a] \\
	\midrule
	\midrule
	CPU & 242.77 & 23.28 & 0.94 & 1.70 \\
	GPU & 226.90 & 45.36 & 1.00 & 1.00 \\
	\midrule
	\multicolumn{5}{l}{FPGA} \\
	\midrule
	25 Mhz & 43.54 & \textbf{14.71} & 5.21 & 83.74 \\
	50 Mhz & 34.95 & 17.53 & 6.49 & 109.06\\
	75 Mhz & 31.96 & 19.02 & 7.10 & 120.24 \\
	100 Mhz & 30.28 & 20.10 & 7.49 & 126.72 \\
	\midrule
	\multicolumn{5}{l}{FPGA + Inference thresholding} \\
	\midrule
	25 Mhz & 35.36 & 17.36 & 6.42 & 107.61 \\
	50 Mhz & 30.81 & 20.11 & 7.36 & 122.35 \\
	75 Mhz & 29.18 & 20.18 & 7.78 & 135.87 \\
	100 Mhz & \textbf{28.53} & 20.53 & \textbf{7.95} & \textbf{139.75} \\
	\bottomrule   
}

Inference thresholding is more beneficial at low operating frequencies.
As the frequency increases, inference time is dominated by the interface between the host and the FPGA.
If this were not the case, we estimate that our approach would use 162 times less energy than the GPU.

Inference thresholding did not have a significant effect on the inference time of our accelerator running on the CPU or GPU.
On the CPU, the output layer only represents a small part of the computation; and the GPU can process the output layer in parallel.

\section{Related Work}


\subsection{DNN Inference Accelerator}
Hardware matrix multiplications can reduce inference times for CNN models~\cite{chen2017eyeriss,han2016eie}.
Several architectures~\cite{han2016eie,han2017ese,rybalkin2017hardware} have been introduced for different types of RNN, such as LSTMs and GRUs.
These accelerators save energy, but are not readily extensible to the memory operations required in MANNs.
A method of accelerating inference of MANNs has been studied~\cite{park2018quantized}, but it has not been implemented in hardware.


\subsection{Maximum Inner-Product Search}
In applications with large search spaces, including NLP, MIPS takes a long time.
Hence, approximations using hashing~\cite{mips-alsh2014}, or clustering~\cite{mips-clustering2015} have been proposed.
Some of these approaches, including sparse access memory~\cite{rae2016scaling} and hierarchical memory networks~\cite{chandar2016hierarchical}, have also been used to accelerate memory reads and writes in MANNs.
However these techniques may be too slow to be used in the output layer of a DNN in resource-limited environments.



\balance
\section{Conclusion}
We believe that the DFA-based approach, and its implementation on an FPGA, which are reported in this paper, represent the first attempt at energy-efficient inference specifically for MANNs.
We also introduce a method of speculation about the inference results which avoids computations which are difficult to parallelize.
This reduces computation times and saves energy at an extremely small cost in accuracy.
We believe that this work shows how inference tasks such as QA may be preformed in mobile devices.
We also expect that our data-based MIPS will find applications in large-class inference.


\section*{Acknowledgements}

This work was supported by the National Research Foundation of Korea (NRF) grant
funded by the Korea government (Ministry of Science and ICT) [2016M3A7B4911115,
2018R1A2B3001628], the Strategic Initiative for Microbiomes in Agriculture and Food (Ministry of Agriculture, Food and Rural Affairs) [918013-4], and the Brain Korea 21 Plus Project in 2018.


\bibliographystyle{IEEEtran}
\bibliography{date}


\end{document}